\def\year{2022}\relax
\documentclass[letterpaper]{article} 
\usepackage{aaai22}  
\usepackage{times}  
\usepackage{helvet}  
\usepackage{courier}  
\usepackage[hyphens]{url}  
\usepackage{graphicx} 
\usepackage{natbib}  
\usepackage{caption} 
\DeclareCaptionStyle{ruled}{labelfont=normalfont,labelsep=colon,strut=off} 
\usepackage{algorithm}
\usepackage{algorithmic}

\usepackage{newfloat}
\usepackage{listings}
\usepackage{xspace,mfirstuc,tabulary}
\usepackage{booktabs}
\usepackage{graphicx}
\usepackage{amssymb}
\usepackage{soul}
\usepackage[normalem]{ulem}
\usepackage{multirow}
\usepackage{amsmath}
\usepackage{array}
\usepackage{enumitem}%
\usepackage{latexsym}
\usepackage{booktabs}
\usepackage{courier}
\usepackage{subfig}
\usepackage{placeins}
\usepackage{tikz}
\usepackage{hyperref}
\usepackage{bbm}

\usepackage{subfiles} 

\floatstyle{ruled}
\newfloat{listing}{tb}{lst}{}
\floatname{listing}{Listing}
\nocopyright

\newcommand{\ra}[1]{\renewcommand{\arraystretch}{#1}}

\title{Hybrid Autoregressive Inference for\\ Scalable Multi-hop Explanation Regeneration}

\author{Marco Valentino$^{1,2}$, Mokanarangan Thayaparan$^{1,2}$, \\Deborah Ferreira$^{1}$, Andr\'e Freitas$^{1,2}$\\}

\affiliations{
Department of Computer Science, University of Manchester, United Kingdom $^{1}$\\ 
Idiap Research Institute, Switzerland $^{2}$
\\ \tt \{name.surname\}@manchester.ac.uk
}

\begin{document}
\maketitle
\begin{abstract}
Regenerating natural language explanations in the scientific domain has been proposed as a benchmark to evaluate complex multi-hop and explainable inference. In this context, large language models can achieve state-of-the-art performance when employed as cross-encoder architectures and fine-tuned on human-annotated explanations. However, while much attention has been devoted to the quality of the explanations, the problem of performing inference efficiently is largely under-studied. Cross-encoders, in fact, are intrinsically not scalable, possessing limited applicability to real-world scenarios that require inference on massive facts banks. To enable complex multi-hop reasoning at scale, this paper focuses on bi-encoder architectures, investigating the problem of scientific explanation regeneration at the intersection of dense and sparse models. Specifically, we present \textbf{SCAR} (for \underline{Sc}alable \underline{A}utoregressive Infe\underline{r}ence), a hybrid framework that iteratively combines a Transformer-based bi-encoder with a sparse model of explanatory power, designed to leverage explicit inference patterns in the explanations. Our experiments demonstrate that the hybrid framework significantly outperforms previous sparse models, achieving performance comparable with that of state-of-the-art cross-encoders while being $\approx 50$ times faster and scalable to corpora of millions of facts. Further analyses on semantic drift and multi-hop question answering reveal that the proposed hybridisation boosts the quality of the most challenging explanations, contributing to improved performance on downstream inference tasks.
\end{abstract}

\section{Introduction}

Explanation regeneration is the task of retrieving and combining two or more facts from an external knowledge source to reconstruct the evidence supporting a certain natural language hypothesis \cite{xie-etal-2020-worldtree,jansen2018worldtree}. As such, this task represents a crucial intermediate step for the development and evaluation of explainable Natural Language Inference (NLI) models \cite{wiegreffe2021teach,thayaparan2020survey}.
In particular, explanation regeneration on science questions has been proposed as a benchmark for complex multi-hop and explainable inference \cite{jansen2019textgraphs,jansen2016s}. Scientific explanations, in fact, require the articulation and integration of commonsense and scientific knowledge for the construction of long explanatory reasoning chains, making multi-hop inference particularly challenging for existing models \cite{clark2018think,khot2019qasc}. Moreover, since the structure of scientific explanations cannot be derived from the decomposition of the questions, the task requires the encoding of complex abstraction and grounding mechanisms for the identification of relevant explanatory knowledge \cite{valentino2021unification,thayaparan-etal-2021-explainable}. 

To tackle these challenges, existing neural approaches leverage the power of the self-attention mechanism in Transformers \cite{devlin2019bert,vaswani2017attention}, training sequence classification models (i.e., cross-encoders) on annotated explanations to compose relevant explanatory chains \cite{cartuyvels-etal-2020-autoregressive,das2019chains,chia2019red,banerjee2019asu}. While Transformers achieve state-of-the-art performance, cross-encoders make multi-hop inference intrinsically inefficient and not scalable to large corpora. The cross-encoder architecture, in fact, does not allow for the construction of dense indexes to cache the encoded explanatory sentences, resulting in prohibitively slow inference time for real-world applications \cite{humeau2019poly}.


\begin{figure*}[t]
\centering
\includegraphics[width=0.9\textwidth]{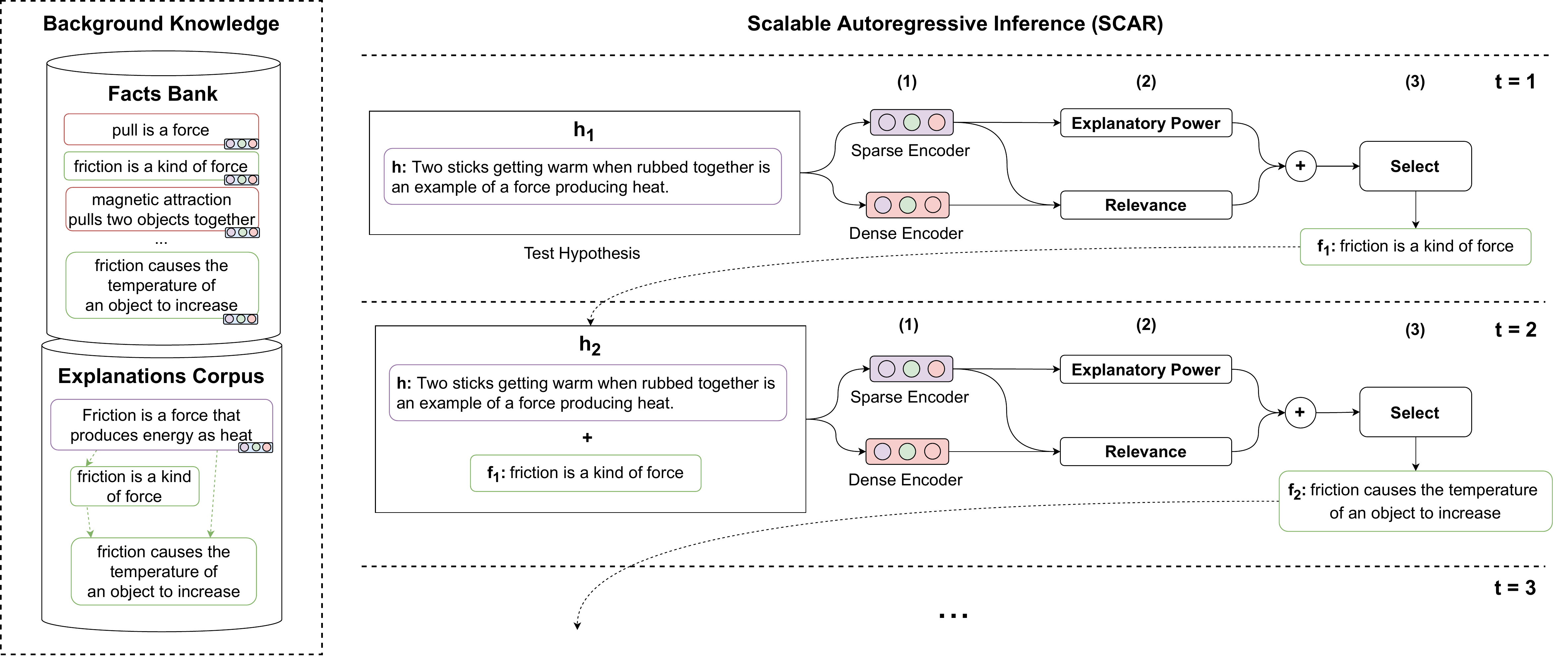}
\caption{We propose a hybrid, scalable explanation regeneration model that performs inference autoregressively. At each time-step $t$, we perform inference integrating sparse and dense bi-encoders (1) to compute relevance and explanatory power of sentences in the Facts Bank (2) and expand the explanation (3). The relevance of a fact at time-step $t$ is conditioned on the partial explanation constructed at time $t-1$, while the explanatory power is estimated leveraging inference patterns emerging across similar hypotheses in the Explanations Corpus.
}
\label{fig:approach}
\end{figure*}

In this work, we are interested in developing new mechanisms to enable scientific explanation regeneration at scale, optimising, at the same time, quality of the explanations and inference time. To this end, we focus our attention on bi-encoders (or siamese networks) \cite{reimers-gurevych-2019-sentence}, which allow for efficient inference via Maximum Inner Product Search (MIPS) \cite{8733051}. Given the complexity of multi-hop reasoning in the scientific domain, bi-encoders are expected to suffer from a drastic drop in performance since the self-attention mechanism cannot be leveraged to learn meaningful compositions of explanatory chains. However, we hypothesise that the orchestration of latent and explicit patterns emerging in natural language explanations can improve the quality of the inference while preserving the scalability intrinsic in bi-encoders.

To validate this hypothesis, we present \textbf{SCAR} (for \underline{Sc}alable \underline{A}utoregressive Infe\underline{r}ence), a hybrid architecture that combines a Transformer-based bi-encoder with a sparse model of explanatory power, designed to capture explicit inference patterns in corpora of scientific explanations.  
Specifically, SCAR integrates sparse and dense encoders to define a joint model of relevance and explanatory power and perform inference in an iterative fashion, conditioning the probability of selecting a fact at time-step $t$ on the partial explanation constructed at time-step $t-1$ (Fig. \ref{fig:approach}).
We performed an extensive evaluation on the WorldTree corpus \cite{jansen2019textgraphs}, presenting the following conclusions: 
\begin{enumerate}
    \item The hybrid framework based on bi-encoders significantly outperforms existing sparse models, achieving performance comparable with that of state-of-the-art cross-encoders while being $\approx 50$ times faster.
    \item  We study the impact of the hybridisation on semantic drift, showing that it makes SCAR more robust in the construction of challenging explanations requiring long reasoning chains.
    \item We investigate the applicability of SCAR on multi-hop question answering without additional training, demonstrating improved accuracy and robustness when performing explainable inference iteratively.
    \item We perform a scalability analysis by gradually expanding the adopted facts bank, showing that SCAR can scale to corpora containing millions of facts.
\end{enumerate}

To the best of our knowledge, we are the first to propose a hybrid autoregressive model for complex multi-hop inference in the scientific domain, demonstrating its efficacy for explanation regeneration at scale.

\section{Multi-hop Explanation Regeneration}

Given a scientific hypothesis $h$ expressed in natural language (e.g., \emph{``Two sticks getting warm when rubbed together is an example of a force producing heat''}), the task of explanation regeneration consists in reconstructing the evidence supporting $h$, composing a sequence of atomic sentences $E_{seq} = f_1,\ldots,f_n$ from external corpora (e.g., \emph{$f_1$:``friction is a kind of force''; $f_2$:``friction causes the temperature of an object to increase''}).
Explanation regeneration can be framed as a multi-hop abductive inference problem, where the goal is to construct the best explanation supporting a given natural language statement adopting multiple retrieval steps. To learn to regenerate scientific explanations, a recent line of research relies on explanation-centred corpora such as Worldtree  \cite{jansen2019textgraphs}, which are typically composed of two distinct knowledge sources (Fig. \ref{fig:approach}):
\begin{enumerate}
    \item A \emph{facts bank} of individual commonsense and scientific sentences including the knowledge necessary to construct explanations for scientific hypotheses.
    \item An \emph{explanations corpus} consisting of true hypotheses and natural language explanations composed of sentences from the facts bank. 
\end{enumerate}


\section{Hybrid Autoregressive Inference}
\label{sec:autoregressive_inference}
To model the multi-hop nature of scientific explanations, we propose a hybrid architecture that performs inference autoregressively (Fig. \ref{fig:approach}). Specifically, we model the probability of composing an explanation sequence $E_{seq} = f_1, \ldots, f_n$ for a certain hypothesis $h$ using the following formulation:

\begin{equation}
\small
P(E_{seq}|h) = \prod_{t=1}^{n}P(f_t|h,f_1, \ldots f_{t-1})
\end{equation}

\noindent where $n$ is the maximum number of inference steps and $f_t$ represents a  fact retrieved at time-step $t$ from the facts bank. We implement the model recursively by updating the hypothesis $h$ at each time-step $t$, concatenating it with the partial explanation constructed at time step $t-1$:

\begin{equation}
\small
h_t = g(h, f_1, \ldots f_{t-1})
\end{equation}

\noindent where $g(\cdot)$ represents the string concatenation function.
The probability $P(f_t|h_t)$ is then approximated via an explanatory scoring function $es(\cdot)$ that jointly models \emph{relevance} and \emph{explanatory power} as:
\begin{equation}
\small
es(f_t, h_t) = \lambda \cdot r(f_t, h_t) + (1-\lambda) \cdot pw(f_t, h)
\label{eq:explanatoryscore}
\end{equation}

\noindent where $r(\cdot)$ represents the relevance of $f_t$ at time step $t$, while $pw(\cdot)$ represents the explanatory power of $f_t$.
As shown in recent work \cite{valentino2021unification}, scientific explanations are composed of abstract sentences describing underlying explanatory laws and regularities that are frequently reused to explain a large set of hypotheses. To leverage this feature during inference, we measure the explanatory power $pw(\cdot)$ of a fact as the extent to which it explains similar hypotheses in the explanations corpus.
The relevance $r(\cdot)$ is computed through a hybrid model that combines a sparse $s(\cdot)$ and a dense $d(\cdot)$ sentence encoder:
\begin{equation}
\small
r(f_t, h_t) = sim(s(f_t), s(h_t)) + sim(d(f_t), d(h_t))
\end{equation}

\noindent with $sim(\cdot)$ representing the cosine similarity between two vectors. In our experiments, we adopt BM25 \cite{robertson2009probabilistic} as a sparse encoder, while Sentence-BERT \cite{reimers-gurevych-2019-sentence} is adopted to train the dense encoder $d(\cdot)$.

\subsection{Explanatory Power}
\label{sec:exp_power}

Recent work have shown that \emph{explicit explanatory patterns} emerge in corpora of natural language explanations \cite{valentino2021unification} -- i.e., facts describing scientific laws and regularities (i.e., laws such as gravity, or friction that explain a large variety of phenomena) are frequently reused to explain similar scientific hypotheses. These patterns can be leveraged to define a computational model of explanatory power for estimating the relevance of abstract scientific facts, providing a framework for efficient explainable inference in the scientific domain.
Given a test hypothesis $h$, a sentence encoder $s(\cdot)$, and a corpus of scientific explanations, the explanatory power of a generic fact $f_i$ can be estimated by analysing explanations for similar hypotheses in the corpus:

\begin{equation}
\small
pw(f_i, h) = \sum_{h_k \in kNN(h)}^K {sim(s(h),s(h_k))} \cdot \mathbbm{1}(f_i, h_k)
\label{eq:unification_score}
\end{equation}  

\begin{equation}
\small
    \mathbbm{1}(f_i,h_k) =
        \begin{cases}
            1 & \text{if } f_i \in E_{k}\\
            0 & \text{if } f_i \notin E_{k}
        \end{cases}
\end{equation}

\noindent where $kNN(h) = \{h_1, \ldots, h_K\}$ represents a list of hypotheses retrieved according to the similarity $sim(\cdot)$ between the embeddings $s(h)$ and $s(h_k)$, and $\mathbbm{1}(\cdot)$ is the indicator function verifying whether $f_i$ is part of the explanation $E_k$ for the hypothesis $h_k$. Specifically, the more a fact $f_i$ is reused for explaining hypotheses that are similar to $h$ in the corpus, the higher its explanatory power.
In this work, we hypothesise that this model can be integrated within a hybrid framework based on dense and sparse encoders, improving inference performance while preserving scalability. In our experiments, we adopt BM25 similarity between hypotheses to compute the explanatory power efficiently.




\begin{table*}[h]
    \small
    \centering
    \ra{1}
    \begin{tabular}{p{7cm}p{8cm}c}
        \toprule
         \multirow{2}{*}{\textbf{Model}} & \multirow{2}{*}{\textbf{Approach Description}} &
         \multirow{2}{*}{\textbf{MAP}}\\\\
       \midrule
       \textbf{Dense Models (Cross-encoders)}\\
       \midrule \citet{cartuyvels-etal-2020-autoregressive} & Autoregressive BERT & \textbf{57.07}\\
       \citet{das2019chains} & BERT path-ranking + single fact ensemble & 56.25\\
       \citet{das2019chains} & BERT single fact & 55.74\\
       \citet{das2019chains} & BERT path-ranking & 53.13\\
       \citet{chia2019red} & BERT re-ranking with gold IR scores & 49.45\\
       \citet{banerjee2019asu} & BERT iterative re-ranking & 41.30\\
         \midrule
       \textbf{Sparse Models}\\
       \midrule
        \citet{valentino2021unification} & Unification-based Reconstruction & \textbf{50.83}\\
        \citet{chia2019red} & Iterative BM25 & 45.76 \\
        BM25 \cite{robertson2009probabilistic} & BM25 Relevance Score & 43.01\\
        TF-IDF & TF-IDF Relevance Score & 39.42\\
        \midrule
        \textbf{Hybrid Models (Bi-encoders)} \\
        \midrule
        SCAR & Scalable Autoregressive Inference & \textbf{56.22} \\
        \bottomrule
    \end{tabular}
    \caption{Results on the test-set and comparison with previous approaches. SCAR significantly outperforms all the sparse models and obtains comparable results with state-of-the-art cross-encoders.}
    \label{tab:compare_approaches_overall}
\end{table*}

\subsection{Dense Bi-encoder}
\label{sec:dense_encoder}
To learn a dense encoder  $d(\cdot)$, we fine-tune a Sentence-BERT model using a bi-encoder architecture \cite{reimers-gurevych-2019-sentence}. The bi-encoder adopts a siamese network to learn a joint embedding space for hypotheses and facts in the facts bank. Following Sentence-BERT, we obtain fixed sized sentence embeddings by adding a mean-pooling operation to the output vectors of BERT \cite{devlin2019bert}. We employ a unique BERT model with shared parameters to learn a sentence encoder $d(\cdot)$ for both facts and hypotheses.
At the cost of sacrificing the performance gain resulting from self-attention, the bi-encoder allows for efficient multi-hop inference through Maximum Inner Product Search (MIPS). To enable scalability, we construct an index of dense embeddings for the whole facts bank. To this end, we adopt the approximated inner product search index (IndexIVFFlat) in FAISS \cite{8733051}.

\paragraph{Training.} The bi-encoder is fine-tuned on inference chains extracted from annotated explanations in the WorldTree corpus \cite{jansen2019textgraphs}. Since the facts in the annotated explanations are not
sorted, to train the model autoregressively, we first transform the explanations into sequences of facts sorting them in decreasing order of BM25 similarity with the hypothesis. We adopt BM25 since the facts that share less terms with
the hypothesis tend to require more iterations and inference steps to be retrieved. Subsequently, given a training hypothesis $h$ and an explanation sequence $E_{seq} = f_1, \ldots, f_n$, we derive $n$ positive example tuples $(h_t, f_t)$, one for each fact $f_t \in E_{seq}$, using $h_t = g(h, f_1, \ldots, f_{t-1})$ as hypothesis. To make the model robust to distracting information, we construct a set of negative examples for each tuple $(h_t, f_t)$ retrieving the top most similar facts to $f_t$ that are not part of the explanation. We found that the best results are obtained using $5$ negative examples for each positive tuple. 
We use the constructed training set and the siamese network to fine-tune the encoder via contrastive loss \cite{contrastive_loss}, which has been demonstrated to be effective for learning robust dense representations.

\subsection{Multi-hop Inference}

At each time-step $t$ during inference time, we encode the concatenation of hypothesis and partial explanation $h_t$ using the dense (Sentence-BERT) and sparse (BM25) encoders separately. Subsequently, we adopt the vectors representing $h_t$ to compute the relevance score $r(\cdot)$ of the sentences in the facts bank (Equation 4). In parallel, the sparse representation (BM25) of the hypothesis $h$ is adopted to retrieve the explanations for the top $K$ similar hypotheses in the explanation corpus and compute the explanatory power $pw(\cdot)$ of each fact (Equation 5). Finally, relevance and explanatory power are combined to compute the explanatory scores $es(\cdot)$ (Equation \ref{eq:explanatoryscore}) and select the top candidate fact $f_t$ from the facts bank to expand the explanation at time-step $t$. After $t_{max}$ steps, we rank the remaining facts considering the explanation constructed a time-step $t_{max}$.

\section{Empirical Evaluation}

We perform an extensive evaluation on the WorldTree corpus adopting the dataset released for the shared task on multi-hop explanation regeneration\footnote{\url{https://github.com/umanlp/tg2019task}} \cite{jansen2019textgraphs}, where a diverse set of sparse and dense models have been evaluated. 
WorldTree is a subset of the ARC corpus \cite{clark2018think} that consists of multiple-choice science questions annotated with natural language explanations supporting the correct answers. The explanations in WorldTree contain an average of six facts and as many as 16, requiring challenging multi-hop inference to be regenerated. The WorldTree corpus provides a held-out test-set consisting of 1,240 science questions with masked explanations where we run the main experiment and comparison with published approaches. To run our experiments, we first transform each question and correct answer pair into a hypothesis following the methodology described in \cite{demszky2018transforming}. We adopt explanations and hypotheses in the training-set ($\approx 1,000$) for training the dense encoder and computing the explanatory power for unseen hypotheses at inference time. 
We adopt \texttt{bert-base-uncased} \cite{devlin2019bert} as a dense encoder to perform a fair comparison with existing cross-encoders employing the same model. The best results on explanation regeneration are obtained when running SCAR for 4 inference steps (additional details in Ablation Studies). In line with the shared task, the performance of the system is evaluated through the Mean Average Precision (MAP) of the produced ranking of facts with respect to the gold explanations in WorldTree. Implementation and pre-trained models adopted for the experiments are available online\footnote{\url{https://github.com/ai-systems/hybrid_autoregressive_inference}}.

\subsection{Explanation Regeneration}

Table~\ref{tab:compare_approaches_overall} reports the results achieved by our best model on the explanation regeneration task together with a comparison with previously published approaches. Specifically, we compare our hybrid framework based on bi-encoders with a variety of sparse and dense retrieval models. 
Overall, we found that SCAR significantly outperforms all the considered sparse models (+5.39 MAP compared to \citet{valentino2021unification}), obtaining, at the same time, comparable results with the state-of-the-art cross-encoder (-0.85 MAP compared to \citet{cartuyvels-etal-2020-autoregressive}). 
The following paragraphs provide a detailed comparison with previous work. 


\paragraph{Dense Models.}

As illustrated in Table~\ref{tab:compare_approaches_overall}, all the considered dense models employ BERT \cite{devlin2019bert} as a cross-encoder architecture. The state-of-the-art model proposed by \citet{cartuyvels-etal-2020-autoregressive} adopts an autoregressive formulation similar to SCAR. However, the use of cross-encoders makes the model computationally expensive and intrinsically not scalable. Due to the complexity of cross-encoders, in fact, the model can only be applied for re-ranking a small set of candidate facts at each iteration, which are retrieved using a pre-filtering step based on TF-IDF.  In contrast, we found that the use of a hybrid model allows achieving comparable performance without cross-attention and pre-filtering step (-0.85 MAP), making SCAR approximately 50 times faster ( see Section \ref{sec:inference_time}).
The second-best dense approach employs an ensemble of two BERT models \cite{das2019chains}. A first BERT model is trained to predict the relevance of each fact individually given a certain hypothesis. A second BERT model is adopted to re-rank a set of two-hops inference chains constructed via TF-IDF. The use of two BERT models in parallel, however, makes the approach computationally exhaustive. We observe that SCAR can achieve similar performance with the use of a single BERT bi-encoder, outperforming each individual sub-component in the ensemble with a drastic improvement in efficiency (SCAR is 96.8 times and 167.4 times faster respectively, see Section \ref{sec:inference_time}).
The remaining dense models \cite{chia2019red,banerjee2019asu} adopt BERT-based cross-encoders to re-rank the list of candidate facts retrieved using sparse Information Retrieval (IR) techniques. As illustrated in Table~\ref{tab:compare_approaches_overall}, SCAR outperforms these approaches by a large margin (+6.77 and +14.92 MAP).

\paragraph{Sparse Models.}

We compare SCAR with sparse models presented on the explanation regeneration task. We observe that SCAR significantly outperforms the Unification-based Reconstruction model proposed by \citet{valentino2021unification} (+5.39 MAP), which employs a model of explanatory power in combination with BM25, but without dense representation and autoregressive inference. These results confirm the contribution of the hybrid model together with the importance of modelling explanation regeneration in a iterative fashion. In addition, we compare SCAR with the model proposed by \citet{chia2019red} which adopts BM25 vectors to retrieve facts iteratively. We found that SCAR can dramatically improve the performance of this model by 10.46 MAP points. Finally, we measure the performance of standalone sparse baselines for a sanity check, showing that SCAR can significantly outperform BM25 and TFIDF (+13.21 and +16.8 MAP respectively), while preserving a similar scalability (see Sec. \ref{sec:scalability}). 
\begin{table}[t]
    \small
    \centering
    \ra{1}
    \begin{tabular}{p{3cm}cc}
        \toprule
         \multirow{2}{*}{\textbf{Model}} & 
         \multirow{2}{*}{\textbf{MAP $\uparrow$}} &
         \multirow{2}{*}{\textbf{Time (s/q) $\downarrow$}}\\\\
        \midrule
        Autoregressive BERT & \textbf{57.07} & 9.6 \\ 
        BERT single fact & 55.74 & 18.4 \\
        BERT path-ranking & 53.13 & 31.8\\
        \midrule
        SCAR & 56.22 (98.5\%) & \textbf{0.19} ($\times50.5$) \\
        \bottomrule
    \end{tabular}
    \caption{Detailed comparison with BERT cross-encoders on the test-set in terms of Mean Average Precision (MAP) and inference time (seconds per question).}
    \label{tab:time_results}
\end{table}

\subsection{Inference Time}
\label{sec:inference_time}

We performed additional experiments to evaluate the efficiency of SCAR and contrast it with state-of-the-art cross-encoders. To this end, we run SCAR on 1 16GB Nvidia Tesla P100 GPU and compare the inference time with that of dense models executed on the same infrastructure \cite{cartuyvels-etal-2020-autoregressive}. Table~\ref{tab:time_results} reports MAP and execution time in terms of seconds per question. 
As evident from the table, we found that SCAR is 50.5 times faster than the state-of-the-art cross-encoder \cite{cartuyvels-etal-2020-autoregressive}, while achieving 98.5\% of its performance. Moreover, when compared to the individual BERT models proposed by \citet{das2019chains}, SCAR is able to achieve better MAP score (+0.48 and +3.09), increasing even more the gap in terms of inference time ($96.8$ and $167.4$ times faster).

\begin{figure*}[t]
\centering
\subfloat[\label{fig:lexical_overlaps}]{\includegraphics[width=0.40\textwidth]{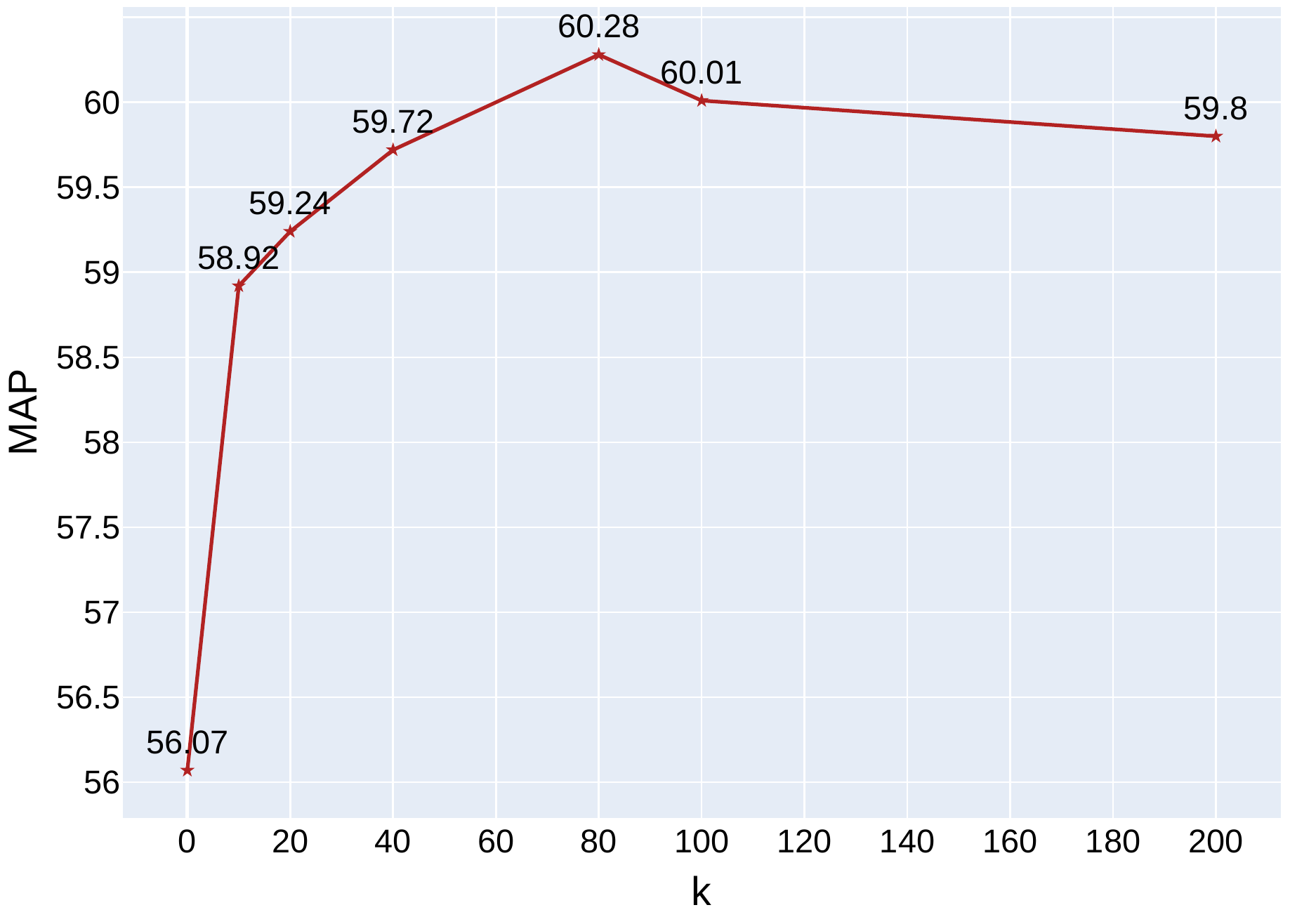}}
\subfloat[\label{fig:map_exp_length}]{\includegraphics[width=0.40\textwidth]{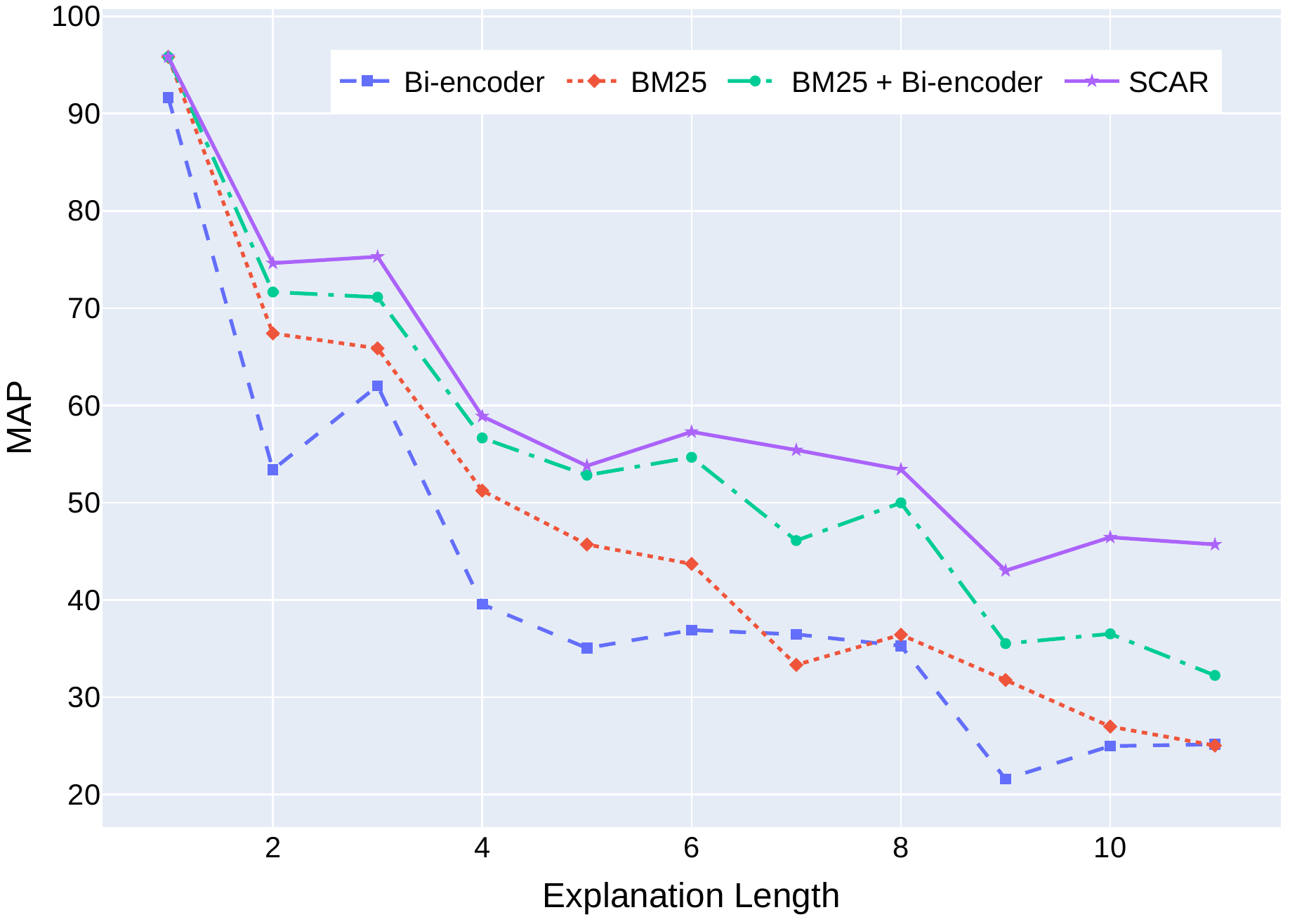}}
\caption{(a) Impact of increasing the number of similar hypotheses $K$ to estimate the explanatory power (Equation \ref{eq:unification_score}). (b) Performance considering hypotheses with gold explanations including an increasing number of facts. 
}
\end{figure*}

\begin{table}[t]
    \small
    \centering
    \ra{1}
    \begin{tabular}{p{3.8cm}ccc}
        \toprule
         \multirow{2}{*}{\textbf{Model}} & \multirow{2}{*}{\textbf{$t_{max}$}} &
         \multirow{2}{*}{\textbf{MAP}} &
         \multirow{2}{*}{\textbf{Time (s/q)}}\\\\
        \midrule
         Bi-encoder & 1 & 41.98 & 0.04 \\ 
         & 2 & \textbf{42.17} & 0.08 \\ 
         & 3 & 39.97 & 0.12 \\
         & 4 & 38.34 & 0.16 \\ 
         & 5 & 37.24 & 0.19 \\ 
         & 6 & 36.64 & 0.24 \\
        \midrule
         BM25 & 1 & 45.99 & 0.02 \\ 
         & 2 & 47.77 & 0.04 \\ 
         & 3 & \textbf{48.35} & 0.05 \\
         & 4 & 48.06 & 0.07\\
         & 5 &  47.97 & 0.09 \\
         & 6 & 47.66 & 0.11\\
        \midrule
        Bi-encoder + BM25 & 1 & 51.53 & 0.05 \\
         & 2 & 54.52 & 0.08 \\
         & 3 & 55.65 & 0.14 \\
         & 4 & 56.07 & 0.18 \\
         & 5 & \textbf{56.24} & 0.22 \\
         & 6 & 55.87 & 0.27 \\
        \midrule
        SCAR & 1 & 57.10 & 0.06 \\
         & 2 & 59.20 & 0.10 \\
         & 3 & 59.73 & 0.15 \\
         & 4 & \underline{\textbf{60.28}} & 0.19 \\
         & 5 & 59.79 & 0.24 \\
         & 6 & 59.36 & 0.29 \\
        \bottomrule
    \end{tabular}
    \caption{Ablation study on the dev-set, where $t_{max}$ represents the maximum number of iterations adopted to regenerate the explanations, and $(s/q)$
    is the inference time.}
    \label{tab:iterative_results}
\end{table}



\subsection{Ablation Studies}
\label{sec:ablation_study}

In order to understand how the different components of SCAR complement each other, we carried out distinct ablation studies. The studies are performed on the dev-set since the explanations on the test-set are masked. 
Table~\ref{tab:iterative_results} presents the results on explanation regeneration for different ablations of SCAR adopting an increasing number of iterations $t_{max}$ for the inference. The results show how the performance improves as we combine sparse and dense models, with a decisive contribution coming from each individual sub-component. Specifically, considering the best results obtained in each case, we observe that SCAR achieves an improvement of 18.11 MAP over the dense component (Bi-encoder) and 11.93 MAP when compared to the sparse model (BM25). Moreover, the ablation demonstrates the fundamental role of the explanatory power model in achieving the final performance, which leads to an improvement of 4.04 MAP over the Bi-encoder + BM25 model (Equation \ref{eq:explanatoryscore}). 
Overall, we notice that performing inference iteratively is beneficial to the performance across the different components. We observe that the improvement is more prominent when comparing $t_{max}=1$ (only using the hypothesis) with $t_{max}=2$ (using hypothesis and first fact), highlighting the central significance of the first retrieved fact to support the complete regeneration process.  Except for the Bi-encoder, the experiments demonstrate a slight improvement when adding more iterations to the process, obtaining the best results for SCAR using a total of 4 inference steps.
We notice that the best performing component in terms of inference time is BM25. The integration with the dense model, in fact, slightly increases the inference time, yet leading to a decisive improvement in terms of MAP score. Even with the overhead caused by the Bi-encoder, however, SCAR can still perform inference in less than half a second per question, a feature that demonstrates the scalability of the approach with respect to the number of iterations.
Finally, we evaluate the impact of the explanatory power model by considering a larger set of training hypotheses for its implementation (Figure~\ref{fig:lexical_overlaps}). To this end, we compare the performance across different configurations with increasing values of $K$ in Equation \ref{eq:unification_score}. The results demonstrate the positive impact of the explanatory power model on the inference, with a rapid increase of MAP peaking at $K=80$. After reaching this value, we observe that considering additional hypotheses in the corpus has little impact on the model's performance. 

\subsection{Semantic Drift}

Recent work have shown that the regeneration of scientific explanations is particularly challenging for multi-hop inference models as it can lead to a phenomenon known as \emph{semantic drift} -- i.e., the composition of spurious inference chains caused by the tendency of drifting away from the original context in the hypothesis \cite{khashabi2019capabilities,xie-etal-2020-worldtree,jansen2019textgraphs,thayaparan2020survey}. In general, the larger the size of the explanation, the higher the probability of semantic drift.
Therefore, it is particularly important to evaluate and compare the robustness of multi-hop inference models on hypotheses requiring long explanations.
To this end, we present a study of semantic drift, comparing the performance of different ablations of SCAR on hypotheses with a varying number of facts in the gold explanations. The results of the study are reported in Figure~\ref{fig:map_exp_length}.
Overall, we observe a degradation in performance for all the considered models that becomes more prominent as the explanations increase in size. Such a degradation is likely due to semantic drift. However, the results suggest that SCAR exhibits more stable performance on long explanations ($\geq$ 6 facts) when compared to its individual sub-components. In particular, the plotted results in Figure \ref{fig:map_exp_length}  clearly show that, while all the models start with comparable MAP scores on explanations containing a single fact, the gap in performance gradually increases with the size of the explanations, with SCAR obtaining an improvement of $13.46$ MAP over BM25 + Bi-encoder on explanations containing more than 10 facts. 
These results confirm the hypotheses that implicit and explicit patterns possess complementary features for explanation regeneration and that the proposed hybridisation has a decisive impact on improving multi-hop inference for scientific hypotheses in the most challenging setting.


\subsection{Multi-hop Question Answering}

 Since the construction of spurious inference chains can lead to wrong answer prediction, semantic drift often influences the downstream capabilities of answering the question. Therefore, we additionally evaluate the performance of SCAR on the multiple-choice question answering task (WorldTree dev-set), employing the model as an explainable solver without additional training.  Specifically, given a multiple-choice science question, we employ SCAR to construct an explanation for each candidate answer, and derive the relative candidate answer score by summing up the explanatory score of each fact in the explanation (Equation \ref{eq:explanatoryscore}). Subsequently, we consider the answer with the highest-scoring explanation as the correct one. Table~\ref{tab:qa_results} shows the results achieved adopting different iterations $t$ for the inference.
Similarly to the results on explanation regeneration, this experiment confirms the interplay between dense and sparse models in improving the performance and robustness on downstream question answering. Specifically, we observe that, while the performance of different ablations decreases rapidly with an increasing number of inference steps, the performance of SCAR are more stable, reaching a peak at $t = 3$. This confirms the robustness of SCAR in multi-hop inference together with its resilience to semantic drift.

\begin{table}[t]
    \small
    \centering
    \ra{1}
    \begin{tabular}{p{3cm}cccc}
        \toprule
         \multirow{2}{*}{\textbf{Model}} & 
         \multirow{2}{*}{\textbf{t = 1}} &
         \multirow{2}{*}{\textbf{t = 2}} &
         \multirow{2}{*}{\textbf{t = 3}} &
         \multirow{2}{*}{\textbf{t = 4}}
         \\\\
        \midrule
        Random & 25.00 & 25.00 & 25.00 & 25.00\\
        \midrule
        BM25 & \textbf{48.23} &  39.82 & 35.84 & 33.18\\
        Bi-encoder & \textbf{54.42} & 52.21 & 50.88 & 50.00\\
        Bi-encoder + BM25 & \textbf{59.29} & 52.21 & 47.79 & 44.69\\ 
        \midrule
        SCAR & 60.62 & 60.62 & \underline{\textbf{61.06}} & 57.96\\
        \bottomrule
    \end{tabular}
    \caption{Accuracy in question answering using the models as explainable inference solvers without additional training.}
    \label{tab:qa_results}
\end{table}

\subsection{Scalability}
\label{sec:scalability}

\begin{figure}[t]
\centering
\includegraphics[width=0.8\columnwidth]{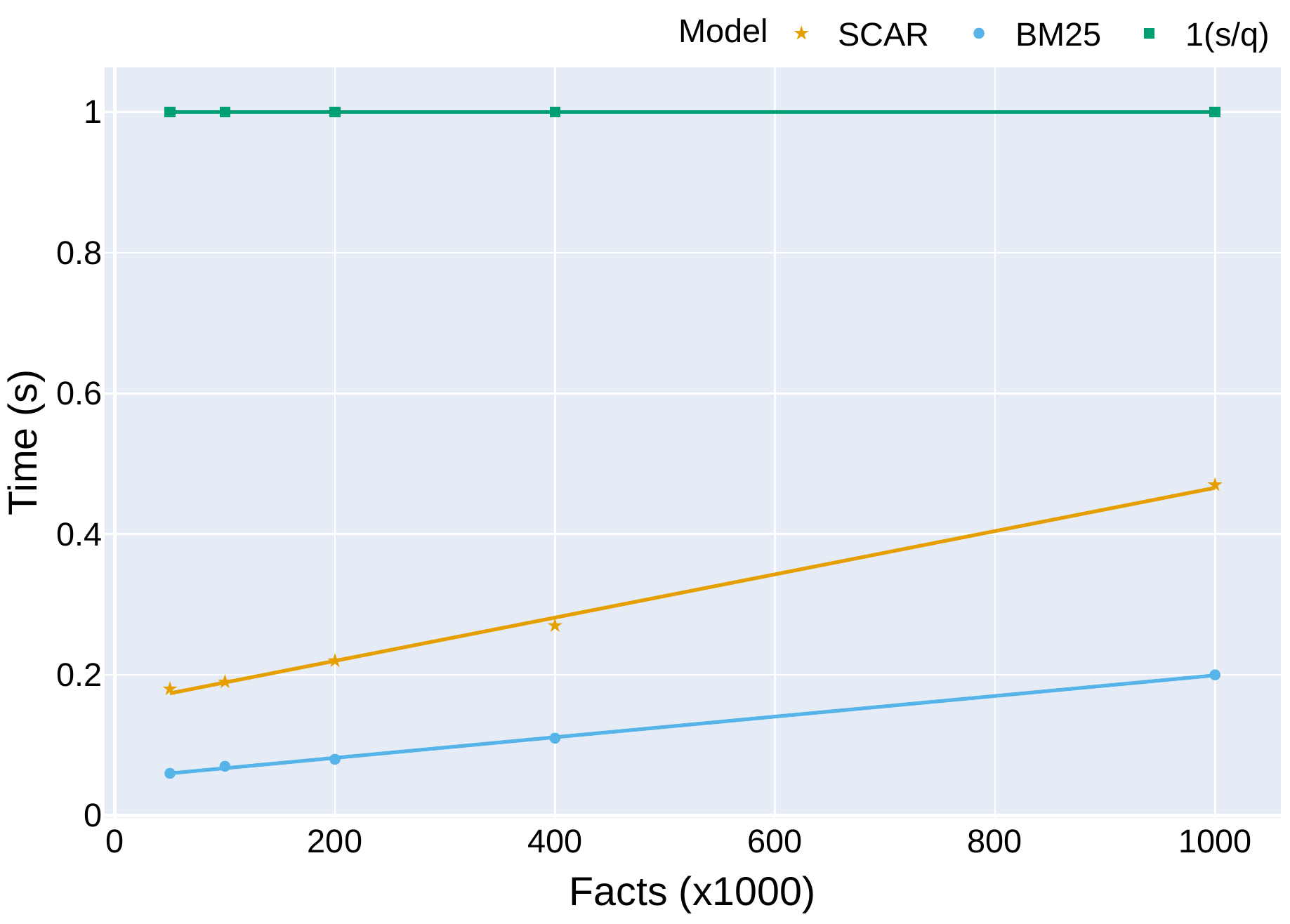}
\caption{Scalability of SCAR to corpora containing a million facts compared to that of standalone BM25.}
\label{fig:scalability_comparison}
\end{figure}

Finally, we measure the scalability of SCAR on facts banks containing milions of sentences. To perform this analysis, we gradually expand the set of facts in the WorldTree corpus by randomly extracting sentences from GenericsKB\footnote{\url{https://allenai.org/data/genericskb}} \cite{bhakthavatsalam2020genericskb}, a curated facts bank of commonsense and scientific knowledge. To evaluate scalability, we compare the inference time of SCAR with that of standalone BM25, which is widely adopted for Information Retrieval at scale \cite{robertson2009probabilistic}. The results of this experiment, reported in Figure \ref{fig:scalability_comparison}, demonstrate that SCAR scales similarly to BM25. Even considering the overhead caused by the Bi-encoder model, in fact, SCAR is still able to perform inference in less than 1 second per question on corpora containing 1 million facts, demonstrating its suitability for real-world scenarios requiring inference on large knowledge sources.

\section{Related Work}

Multi-hop inference is the task of combining multiple pieces of evidence to solve a particular reasoning problem. This task is often used to evaluate explainable inference since the constructed chains of reasoning can be interpreted as an explanation for the final predictions \cite{wiegreffe2021teach,thayaparan2020survey}.
Given the importance of multi-hop reasoning for explainability, there is a recent focus on resources providing annotated explanations to support the inference \cite{xie-etal-2020-worldtree,jhamtani2020learning,Khot_Clark_Guerquin_Jansen_Sabharwal_2020,ferreira2020natural,khashabi2018looking,yang2018hotpotqa,mihaylov2018can,jansen2018worldtree,welbl2018constructing}. 
In particular, explanation regeneration on science questions is designed to evaluate the construction of long explanatory chains, in a setting where the structure of the inference cannot be derived from a direct decomposition of the questions \cite{xie-etal-2020-worldtree,jansen2019textgraphs,jansen2018worldtree}.
To deal with the difficulty of the task, state-of-the-art models leverage the attention mechanism in Transformers \cite{vaswani2017attention}, learning to compose relevant explanatory chains via sequence classification models \cite{cartuyvels-etal-2020-autoregressive,das2019chains,chia2019red,banerjee2019asu}.
The autoregressive formulation proposed in this paper is similar to the one introduced by \citet{cartuyvels-etal-2020-autoregressive}, which, however, perform iterative inference though a cross-encoder architecture based on BERT \cite{devlin2019bert}. Differently from this work, we present a hybrid architecture based on bi-encoders \cite{reimers-gurevych-2019-sentence} with the aim of optimising both accuracy and inference time in explanation regeneration. Our model of explanatory power is based on the work done by \citet{valentino2021unification}, which shows how to leverage explicit patterns in corpora of scientific explanations for multi-hop inference. In this paper, we build upon this line of research by demonstrating that models based on explicit patterns can be combined with neural architectures to achieve nearly state-of-the-art performance while preserving scalability. 
Our framework is related to recent work on dense retrieval for knowledge-intensive NLP tasks, which focuses on the design of scalable architectures with Maximum Inner Product Search (MIPS) based on Transformers \cite{xiong2021answering,zhao-etal-2021-multi-step,lin-etal-2021-differentiable,karpukhin-etal-2020-dense,NEURIPS2020_6b493230,dhingra2019differentiable}. Our multi-hop dense encoder is similar to \citet{lin-etal-2021-differentiable} and \citet{xiong2021answering} which adopt bi-encoders for multi-step retrieval on open-ended commonsense reasoning and open-domain question answering. However, to the best of our knowledge, we are the first to integrate dense bi-encoders in a hybrid architecture for complex explainable inference in the scientific domain. 

\section{Conclusion}

This work presented SCAR, a hybrid autoregressive architecture for scalable explanation regeneration.
An extensive evaluation demonstrated that SCAR achieves performance comparable with that of state-of-the-art cross-encoders while being $\approx 50$ times faster and intrinsically scalable, and confirmed the impact of the hybridisation on semantic drift and question answering.
This work demonstrated the effectiveness of hybrid architectures for explainable inference at scale, opening the way for future research at the intersection of latent and explicit models. As a future work, we plan to investigate the integration of relevance and explanatory power in an end-to-end differentiable architecture, and explore the applicability of the hybrid framework on additional natural language and scientific reasoning tasks, with a focus on real-world scientific inference problems. 

\bibliography{acl}
\appendix
    \section{Hyperparameters tuning}
\begin{figure}[t]
    \centering
    \includegraphics[width=0.8\columnwidth]{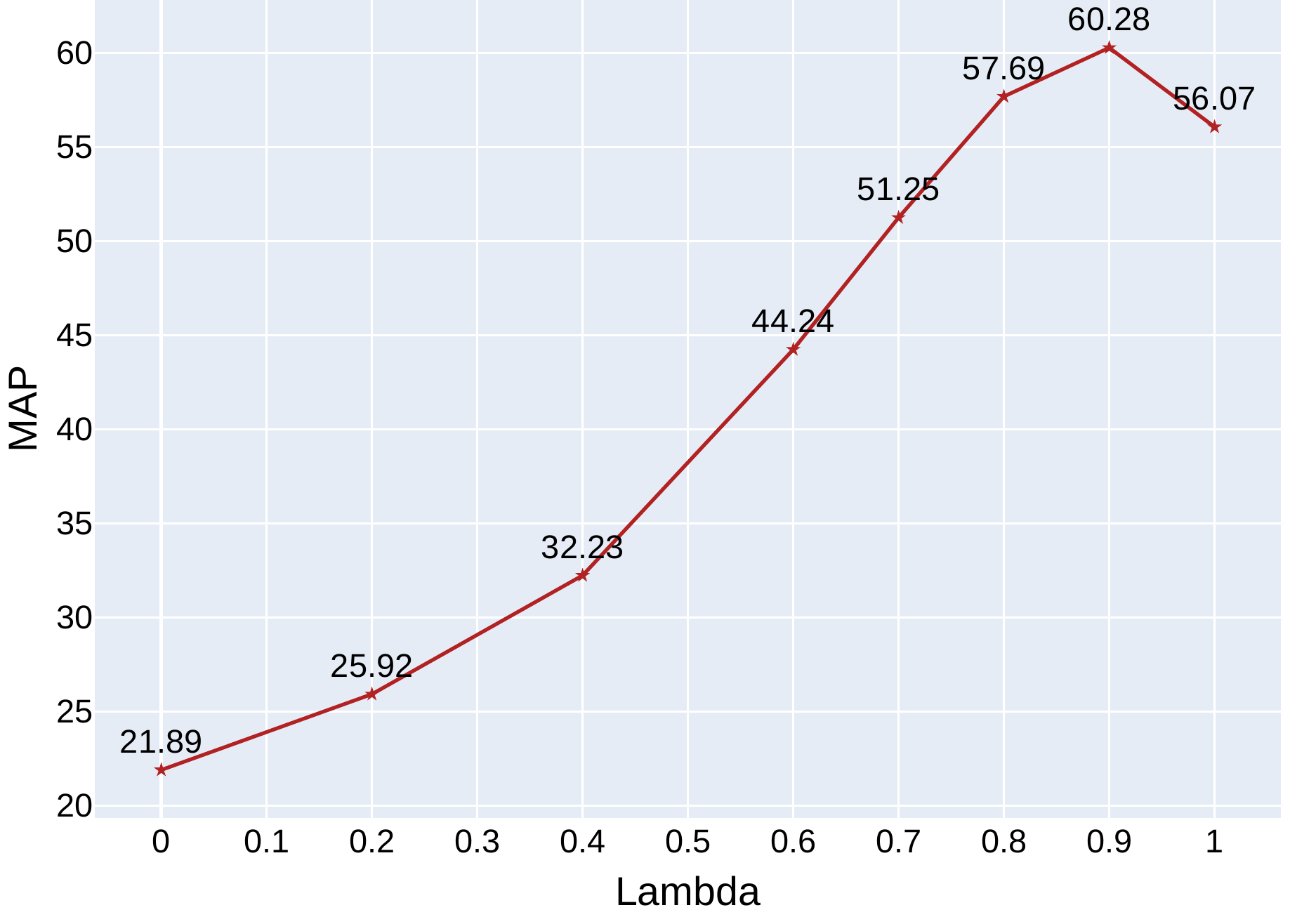}
    \caption{Tuning of $\lambda$ for the explanatory scoring function $es(\cdot)$ (Equation 3) on the dev-set.}
    \label{fig:fine_tuning}
    \end{figure}

    Tthe hyperparameters of SCAR have been tuned to maximise the MAP for explanation regeneration on the WorldTree dev-set. Here, we report the best values for $\lambda$, which is used to assign the weights to relevance and explanatory power for computing the explanatory scoring function described in Equation 3:
    \begin{itemize}
        \item $\lambda$ = 0.89
    \end{itemize}
    Figure \ref{fig:fine_tuning} shows the MAP score obtained with different values of $\lambda$ with 0 representing the extreme case in which only the explanatory power is active and 1 the case in which only the relevance is active. As shown in the graph, the explanatory power alone does not allow to achieve high performance on the task, demonstrating that explanations for unseen  hypotheses cannot be simply regenerated considering similar hypotheses in the training set and that the relevance model is necessary for generalisation on unseen examples.
    
    \section{Dense Encoder}
    For the implementation of the dense encoder $d(\cdot)$ we adopt Sentence-BERT, whose package can be found at the following URL: \url{https://pypi.org/project/sentence-transformers/}. Specifically, we implement the bi-encoder with a \emph{bert-base-uncased} model, adopting a mean-pooling operation to obtain fixed sized sentence embeddings and contrastive loss for training. We release the trained model adopted in the experiments at the following URL: \url{https://drive.google.com/file/d/1iz38q8EIYZdO9U7mAMVz1qUprU8jmEwI/view}. 
    
    \subsection{Training Setup}
    We train the model using 1 16GB Nvidia Tesla P100 GPU for 3 epochs in total with contrastive loss, while 10\% of the training data is used for warm-up.
    We adopted the following hyperparameters for training:
    \begin{itemize}
        \item batch size = 16
        \item margin (contrastive loss) = 0.25
        \item learning\_rate = 2e-5
        \item weight\_decay = 0.1
        \item adam\_epsilon = 1e-8
        \item max\_grad\_norm = 1.0
    \end{itemize}
    
    \subsection{Faiss Index}
    
    For creating the index of dense vectors for the facts bank we use the Faiss package for Python available at the following URL: \url{https://pypi.org/project/faiss-gpu/}. Specifically, we adopt IndexIVFFlat.
    
    \section{Source Code and Data}
    The complete code adopted to run our experiments is available at the following URL: \url{https://github.com/ai-systems/hybrid_autoregressive_inference}.
    The WorldTree corpus can be downloaded at the following url: \url{http://cognitiveai.org/dist/worldtree_corpus_textgraphs2019sharedtask_withgraphvis.zip}.

\end{document}